\documentclass[11pt]{article}

\usepackage[a4paper,margin=1in]{geometry}
\usepackage{amsmath}
\usepackage{mathptmx}
\usepackage[scaled=0.92]{helvet}
\usepackage{microtype}
\usepackage[dvipsnames]{xcolor}
\usepackage{tikz}
\usetikzlibrary{positioning,fit,backgrounds,arrows.meta,calc,shapes.geometric,shadows}
\usepackage{booktabs}
\usepackage{array}
\usepackage{enumitem}
\usepackage[hidelinks]{hyperref}
\hypersetup{pdftitle={CrossAudit: A Git-Native, Cross-Vendor Audit Loop for Agentic Science},
  pdfauthor={Zhaohe Dong, Yuhao Chen},
  pdfkeywords={agentic science, AI scientist, scalable oversight, cross-vendor auditing, LLM-as-a-judge, self-preference bias, research integrity, human-in-the-loop}}
\usepackage{caption}
\definecolor{caindigo}{HTML}{4F46E5}
\definecolor{cagreen}{HTML}{059669}
\definecolor{caorange}{HTML}{EA580C}
\definecolor{caslate}{HTML}{475569}
\definecolor{cared}{HTML}{DC2626}
\definecolor{caviolet}{HTML}{8B5CF6}
\definecolor{cabgindigo}{HTML}{EEF2FF}
\definecolor{cabggreen}{HTML}{ECFDF5}
\definecolor{cabgorange}{HTML}{FFF7ED}
\definecolor{cabgslate}{HTML}{F8FAFC}
\definecolor{cabgred}{HTML}{FEF2F2}
\definecolor{cabgviolet}{HTML}{F5F3FF}

\newcommand{\invariant}[1]{\textbf{I#1}}
\newcommand{\rulename}[1]{\texttt{#1}}
\newcommand{\tnotea}{$^{a}$}\newcommand{\tnoteb}{$^{b}$}\newcommand{\tnotec}{$^{c}$}

\title{\textbf{CrossAudit: A Git-Native, Cross-Vendor Audit Loop\\for Agentic Science}}
\author{Zhaohe Dong\\
\normalsize University of Cambridge\\
\normalsize \texttt{zd314@cam.ac.uk}
\and Yuhao Chen\\
\normalsize University of Wisconsin--Madison\\
\normalsize \texttt{ychen2546@wisc.edu}}
\date{August 2026}

\begin{document}
\maketitle

\begin{abstract}
\noindent An AI scientist should not grade its own homework. Yet in the systems we examined, the agent that reviews the work usually comes from the same model family as the agent that produced it, or at least from the same vendor. Model evaluators are known to favour their own generations. Whether models trained alike also share blind spots is a conjecture, not a settled finding, but if they do, the reviewer inherits the author's. The record of what was flagged and what was waved through often sits in platform logs that nobody outside can replay.

We present \textbf{CrossAudit}, a protocol for supervising autonomous research pipelines. It rests on three commitments. Each increment of work is audited by an agent from a \emph{different} vendor, against a rulebook a human wrote and versioned. Reports, verdicts, disputes and rulings are git commits, so the supervision history can be re-read and cited; raw model exchanges are not yet part of that record. Scripted checks run before any model does. Advisory judgement never gates the pipeline: a model blocks only by citing a rule, and no model may waive a deterministic failure. Blockers that survive a bounded number of revision rounds go to a person.

We state the protocol as eight invariants. We describe a reference implementation built from GitHub Actions and a few hundred lines of Python, and report a live deployment of a closely related variant in a computational-chemistry pipeline. We also ran a seeded-defect trial (30 increments, 43 seeded defects, one run per configuration). A cross-vendor audit of our own repository then voided its blinding. We adopt that audit's findings and report the corrected results. The trial shows that two vendors read the same rulebook differently. It does not show that either is better. The strongest evidence here is the committed, uncontrolled record of cross-vendor audits of this paper itself.
\end{abstract}

\vspace{5pt}\noindent{\small\textbf{Keywords:} agentic science; AI scientist; scalable oversight; cross-vendor auditing; LLM-as-a-judge; self-preference bias; research integrity; human-in-the-loop.}

\section{Introduction}\label{sec:intro}

Two things motivated this paper. The first is the arrival of genuinely agentic research systems: pipelines in which large language model (LLM) agents propose hypotheses, write and execute code, run experiments, and draft manuscripts with limited human intervention. Sakana's AI Scientist produced machine-generated papers end-to-end \cite{lu2024aiscientist,yamada2025aisv2} and the line of work has since been published in \emph{Nature} \cite{lu2026nature}; DeepMind's AI co-scientist organises hypothesis generation as a multi-agent tournament \cite{gottweis2025coscientist}; autonomous laboratories close the loop through to physical synthesis \cite{boiko2023,szymanski2023}. Publishers and funders are already being urged to respond institutionally \cite{naturenews2026}.

The second is older and more familiar: science's continuing struggle with reliability. In one large survey, most researchers had failed to reproduce a published result \cite{baker2016}. Agentic pipelines raise the stakes on both sides of this ledger. On one side, they can enforce a discipline that no human sustains: every increment logged, every parameter recorded. On the other, they can mass-produce plausible, well-formatted, wrong results faster than any human audit culture can absorb.

The relevant question, then, is not whether machine-generated science should be supervised, but how that supervision should be organised and who should provide it. Current frontier systems typically place supervision within the same architecture, using reflection agents, automated reviewers, or ranking tournaments \cite{lu2024aiscientist,gottweis2025coscientist}. This creates a structural weakness: the critic and the creator are usually instantiated from the same model family or, at best, from models developed within the same vendor's training pipeline. LLM evaluators are known to favour their own generations, with self-recognition identified as a causal factor in this self-preference bias \cite{panickssery2024}; position and stylistic biases introduce further distortions \cite{zheng2023}. Agents trained on similar data may also share systematic blind spots, so a reviewer with the same blind spots as the author may fail to detect the author's characteristic errors. At the same time, the supervision record (what was flagged, what was revised, and what was allowed to pass) often remains confined to transient context windows or proprietary platform logs. As a result, the communities asked to trust the resulting science cannot independently inspect the process by which it was reviewed.

\paragraph{Contributions.} We make four contributions.
\begin{enumerate}[itemsep=1pt,topsep=3pt]
  \item We articulate the \emph{same-source supervision problem} in agentic science: internal critique by same-vendor models is structurally exposed to correlated failure (\S\ref{sec:related}).
  \item We specify \textbf{CrossAudit}, a supervision protocol defined by eight invariants (vendor heterogeneity, ledger completeness, citation validity, determinism-first, bounded revision, graded interruption, receipt binding, fail-closed admission) that together specify a re-inspectable, third-party-auditable (up to self-declared vendor identity, \S\ref{sec:threat}) supervision history (\S\ref{sec:protocol}).
  \item We describe a reference implementation requiring no infrastructure beyond two git repositories, a CI service, and two model API keys, together with a live deployment of a closely related variant in the first author's computational-chemistry pipeline, and we report what a second, product-oriented implementation of the same protocol taught us about the specification (\S\ref{sec:implementation}).
  \item We give an explicit threat model, including the attacks the protocol does \emph{not} withstand, and situate the design against plausible alternatives (\S\ref{sec:threat}--\S\ref{sec:discussion}).
\end{enumerate}

CrossAudit is deliberately limited in what it claims to establish. It does not certify scientific truth. No textual audit can. Instead, it records a narrower and directly inspectable property: that each increment in a machine-generated research record was evaluated against a named, versioned set of rules by a configured auditor from a second vendor, and that the resulting parsed decision was preserved in the ledger. The auditor has no write access to the increment's repository, although its vendor identity is currently self-declared (\S\ref{sec:threat}). The ledger is public wherever the operator chooses to publish it. At present it preserves parsed audit reports, while capture of the raw model requests and responses is planned for revision~2. We argue that this narrower guarantee is an appropriate first target for accountable machine science, much as continuous integration, not formal verification, became the principal quality-control mechanism in modern software development.

\section{Related work and the same-source problem}\label{sec:related}

\paragraph{Agentic science systems.} The AI Scientist \cite{lu2024aiscientist,yamada2025aisv2,lu2026nature} automates the ideation--experiment--writeup--review cycle, including an internal automated reviewer scored against human reviewer data. The AI co-scientist \cite{gottweis2025coscientist} structures generation--reflection--ranking--evolution agents into a self-improving tournament over hypotheses, with reported expert-validated outcomes in drug repurposing and antimicrobial resistance. Language agents for literature synthesis reach or exceed expert performance on retrieval-grounded tasks \cite{skarlinski2024}.  In the laboratory, Coscientist executed chemistry protocols from natural-language goals \cite{boiko2023} and A-Lab ran autonomous materials synthesis campaigns \cite{szymanski2023}. Across all of these systems, quality control is architecturally \emph{internal}. The reviewer agents, reflection stages, or tournament judges are chosen and hosted by the same operator, run inside the same platform, and are frequently instantiated from the same model family as the generators they judge. None of them commits to vendor heterogeneity in the reviewing layer.

The pattern persists through the 2025 generation of research agents: Curie hardens agentic experimentation with internal rigour modules \cite{kon2025curie}, and Agent Laboratory attaches critic agents to its own pipeline \cite{schmidgall2025agentlab}. Commercial platforms now advertise the transparency half of our proposal: Kosmos promises reports in which ``every conclusion can be traced'' to code or literature via its platform \cite{edison2025kosmos}. Traceability of that kind is audit by the generating operator on its own infrastructure. It is valuable, but it involves no second-party auditor and no off-platform, replayable ledger. This paper is about that distinction.

\paragraph{Why internal critique is not enough.} Panickssery et al.\ demonstrate that LLM evaluators recognise their own generations and rate them preferentially. Their fine-tuning experiments indicate that the capacity for self-recognition is what causes the inflation \cite{panickssery2024}. Zheng et al.\ catalogue further judge pathologies, including position, verbosity, and self-enhancement biases, even in strong models \cite{zheng2023}. A recent large-scale evaluation sharpens why these matter for supervision rather than for benchmarking: in a twenty-one-judge cohort, two production-deployed judges combine high test-retest reliability with severe position bias, and raw agreement overstates judge quality substantially once corrected for chance \cite{norman2026reliability}. A judge can therefore be thoroughly reproducible and still be wrong in a fixed direction, which is the one failure a supervision loop must not inherit. Two later studies qualify the self-preference finding rather than overturn it: much of the measured preference tracks genuine quality differences between generations \cite{chen2025selfpref}, and identity-blind re-tests leave roughly half of previously significant cases standing \cite{roytburg2026sanity}. The configuration CrossAudit forecloses is therefore a narrower target than the headline effect, and we treat foreclosing it as a structural precaution, not a quantified harm averted. These results concern single-model self-evaluation. We expect the mechanism to generalise. Models that share a training pipeline are likely to share stylistic priors and knowledge gaps, and this places a same-vendor critic at the wrong end of the bias distribution just when its judgement matters most. We call this the \emph{same-source supervision problem}. The remedy of drawing judges from different model families has precedent. Verga et al.\ score generations with a panel of evaluators drawn from three model families, expressly to dilute intra-model bias \cite{verga2024poll}, and the same intuition has begun to appear in developer tooling as cross-vendor code review.\footnote{E.g.\ MindStudio, ``Cross-Vendor AI Agent Review: Why Claude Should Review Codex's Code and Vice Versa'' (June 2026), \url{https://www.mindstudio.ai/blog/cross-vendor-ai-agent-review-claude-codex}, as accessed 30 July 2026, a workflow recommendation, without versioned rules, ledger, deterministic precedence, or escalation semantics.} CrossAudit's claim is not the heterogeneity idea itself but its packaging as a supervision protocol for research: versioned rules, a replayable ledger, deterministic precedence, and bounded escalation, run against a live pipeline. Cross-vendor pairing does not eliminate correlated error. Frontier models plausibly train on largely overlapping public corpora, so where the shared literature is wrong, both auditors can be confidently wrong together. What pairing does foreclose is the demonstrated configuration of the best-documented bias in this setting, self-preference, and it is intended to decorrelate vendor-idiosyncratic failure modes. What cross-vendor pairing cannot fix, a model-free layer must (\S\ref{sec:protocol}, \invariant{4}).

\paragraph{Supervision as an artefact.} Software engineering solved an analogous trust problem not by making programmers infallible but by making the quality mechanisms \emph{inspectable}: version control, continuous integration, review trails. Reproducibility initiatives in science point in the same direction. The unit of trust is the auditable record, not the author's competence \cite{baker2016}. Deterministic screening of the research record also has its own lineage. statcheck recomputes reported statistics across the psychology literature \cite{nuijten2016statcheck}, and the Problematic Paper Screener runs mechanical detectors over millions of published papers \cite{cabanac2022pps}. Both are post-publication analogues of our deterministic check layer, which moves checks of that character upstream, before a claim ever leaves the repository. At the ledger end, blockchain-based research-integrity infrastructures pursue tamper-evidence with far heavier machinery.\footnote{E.g.\ the bloxberg consortium, \url{https://bloxberg.org}.} Our position is that git, which is ubiquitous, free, and already attached to CI, permissions, and issue tracking, is ledger enough for the supervision trail. CrossAudit transplants this inspectability stance to agentic science: the supervision history itself becomes a first-class, versioned, citable research artefact. Table~\ref{tab:related} compresses the comparison.


\begin{table}[t]
\centering\footnotesize\setlength{\tabcolsep}{3.5pt}
\caption{Supervision properties of the systems considered here (a set we chose, not a systematic survey), as published. $\bullet$ = held by design; $\sim$ = partial or informal; -- = absent. PoLL is an evaluation technique, not a supervision loop; statcheck and the Problematic Paper Screener are deterministic by construction but run post-publication, where CrossAudit moves checks of that character before a claim leaves the repository. CrossAudit appears twice so that a specification is never graded on an implementation's behalf: the protocol row states what the specification requires, and the implementation row grades the public reference implementation, whose ledger preserves parsed records but not raw exchanges (\S\ref{sec:implementation}) and whose auditor vendor identity is self-declared (\S\ref{sec:threat}). The ledger column is graded at \invariant{2}'s parsed-record tier; full process replay is unachievable while model calls are stochastic, for the specification as much as for any implementation.}
\label{tab:related}
\begin{tabular}{@{}lccccc@{}}
\toprule
 & \begin{tabular}{@{}c@{}}Critic from a\\different vendor\end{tabular}
 & \begin{tabular}{@{}c@{}}Versioned,\\decidable rules\end{tabular}
 & \begin{tabular}{@{}c@{}}Replayable\\audit ledger\end{tabular}
 & \begin{tabular}{@{}c@{}}Deterministic\\precedence\end{tabular}
 & \begin{tabular}{@{}c@{}}Bounded loop,\\escalation\end{tabular} \\
\midrule
The AI Scientist \cite{lu2024aiscientist} & -- & -- & $\sim$ & -- & $\sim$ \\
AI co-scientist \cite{gottweis2025coscientist} & -- & -- & -- & -- & $\sim$ \\
PoLL judge panels \cite{verga2024poll} & $\bullet$ & -- & -- & -- & -- \\
Cross-vendor review advice (\S\ref{sec:related}) & $\bullet$ & -- & $\sim$ & -- & -- \\
statcheck, PPS \cite{nuijten2016statcheck,cabanac2022pps} & -- & $\sim$ & $\sim$ & $\bullet$ & -- \\
\textbf{CrossAudit protocol (this paper)} & $\bullet$ & $\bullet$ & $\bullet$ & $\bullet$ & $\bullet$ \\
\textbf{CrossAudit reference implementation} & $\sim$ & $\bullet$ & $\sim$ & $\bullet$ & $\bullet$ \\
\bottomrule
\end{tabular}
\end{table}

\section{The CrossAudit protocol}\label{sec:protocol}

\subsection{Objects and roles}\label{sec:roles}

CrossAudit supervises a pipeline in which a \textbf{Generator} agent (vendor $A$) produces discrete \emph{experiment increments} and commits them under a designated path of a \textbf{Science Repository}. Each increment is audited by an \textbf{Auditor} agent (vendor $B \neq A$) against a versioned, human-authored rulebook, the \textbf{Constitution}, held with all \emph{audit reports} in a separate \textbf{Audit Repository}. A \textbf{Deterministic Check Layer} (DCL), plain scripts with no learned component (schema validation, unit and dimension analysis, convergence assertions, provenance completeness), runs before any model reads anything. A \textbf{Human Principal} authors the Constitution and receives escalations, but sits outside the per-increment loop. Compute (in the reference deployment, jobs on cloud HPC) is decoupled from supervision: verdicts gate the \emph{next} increment, never running jobs. Figure~\ref{fig:arch} traces one increment through the loop.

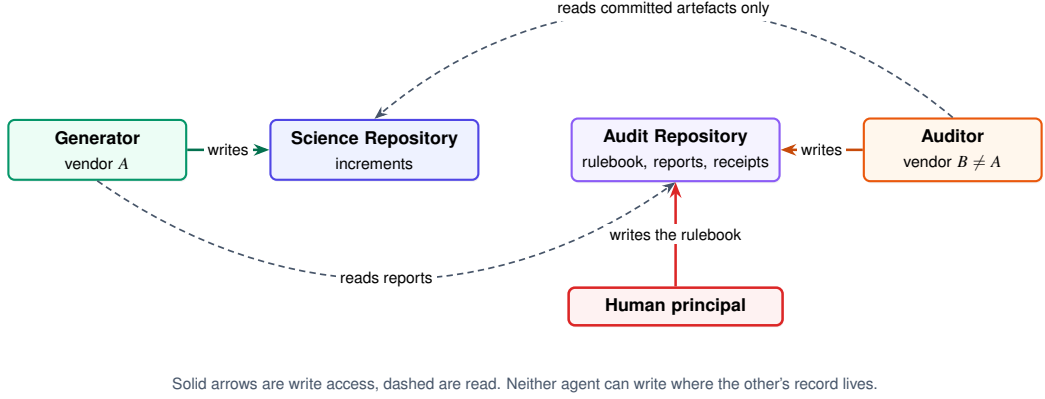
\begin{figure}[t]
\centering
\resizebox{0.86\textwidth}{!}{%
\begin{tikzpicture}[x=1cm, y=1cm, font=\footnotesize\sffamily,
  every shadow/.style={shadow xshift=0.5pt, shadow yshift=-1pt, opacity=0.10},
  box/.style={draw, rounded corners=3pt, align=center, inner sep=5pt, line width=1pt, drop shadow},
  ag/.style={box, text width=26mm}, rp/.style={box, text width=31mm},
  w/.style={-{Stealth[length=2.6mm]}, line width=1.2pt},
  rd/.style={-{Stealth[length=2.2mm]}, line width=0.8pt, densely dashed, draw=caslate},
  lbl/.style={font=\scriptsize\sffamily, fill=white, inner sep=1.8pt}]
\node[ag, draw=cagreen, fill=cabggreen] (gen) at (0,0) {\textbf{Generator}\\[1pt] \scriptsize vendor $A$};
\node[rp, draw=caindigo, fill=cabgindigo] (sci) at (4.6,0) {\textbf{Science Repository}\\[1pt] \scriptsize increments};
\node[rp, draw=caviolet, fill=cabgviolet] (led) at (9.6,0) {\textbf{Audit Repository}\\[1pt] \scriptsize rulebook, reports, receipts};
\node[ag, draw=caorange, fill=cabgorange] (aud) at (14.2,0) {\textbf{Auditor}\\[1pt] \scriptsize vendor $B \neq A$};
\node[box, draw=cared, fill=cabgred, text width=32mm, line width=1.2pt] (hum) at (9.6,-2.6) {\textbf{Human principal}};
\draw[w, draw=cagreen!75!black] (gen) -- node[lbl]{writes} (sci);
\draw[w, draw=caorange!85!black] (aud) -- node[lbl]{writes} (led);
\draw[rd] (aud.north) to[out=140, in=40] node[lbl]{reads committed artefacts only} (sci.north);
\draw[rd] (gen.south) to[out=-35, in=-145] node[lbl]{reads reports} (led.south);
\draw[w, draw=cared] (hum) -- node[lbl]{writes the rulebook} (led);
\node[font=\scriptsize\sffamily, text=caslate, align=center] at (7.1,-3.9)
  {Solid arrows are write access, dashed are read. Neither agent can write where the other's record lives.};
\end{tikzpicture}}
\caption{Separation of powers. The Generator cannot touch the rules or the reports that judge it, and the Auditor cannot touch the record it audits. The Auditor reads committed artefacts and never the Generator's prompts or reasoning. Only the human principal writes the rulebook. Guarding each record against its \emph{own} author also requires branch protection.}
\label{fig:powers}
\end{figure}

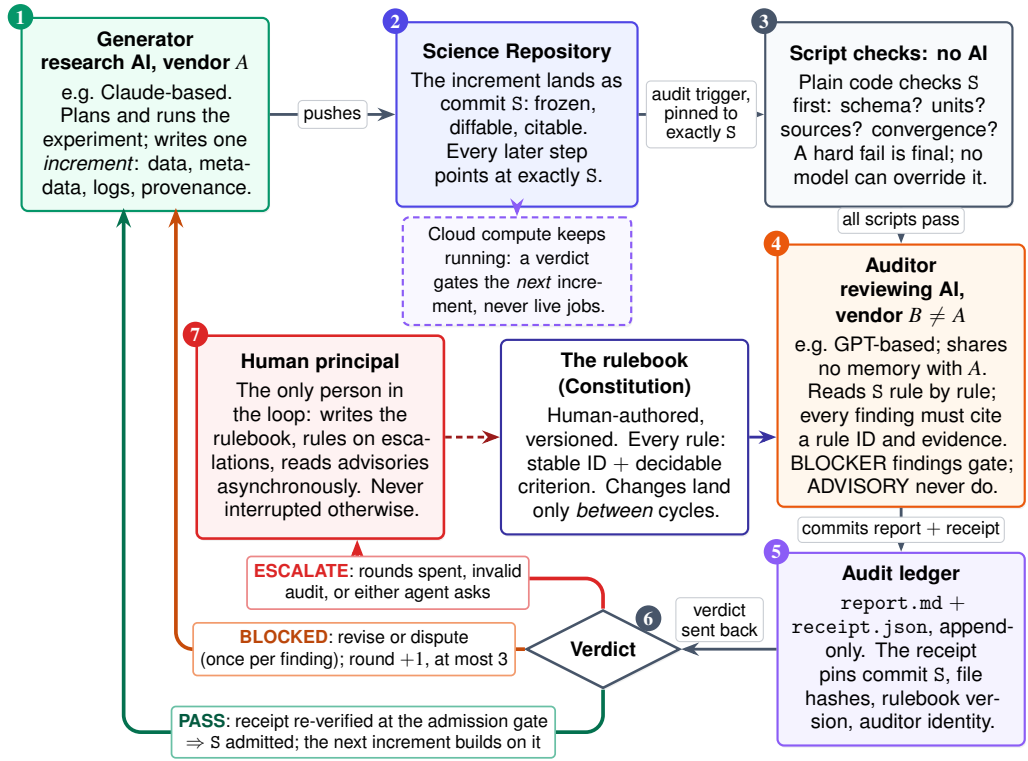
\begin{figure}[!tp]
\centering
\resizebox{\textwidth}{!}{%
\begin{tikzpicture}[x=1cm, y=1cm,
  font=\footnotesize\sffamily,
  every shadow/.style={shadow xshift=0.6pt, shadow yshift=-1.1pt, opacity=0.12},
  box/.style={draw, rounded corners=3pt, align=center, inner sep=6pt, line width=1pt, drop shadow},
  gen/.style={box, draw=cagreen, fill=cabggreen, text width=36mm, minimum height=30mm},
  sci/.style={box, draw=caindigo, fill=cabgindigo, text width=35mm, minimum height=30mm},
  det/.style={box, draw=caslate, fill=cabgslate, text width=36mm, minimum height=30mm},
  aud/.style={box, draw=caorange, fill=cabgorange, text width=36mm},
  led/.style={box, draw=caviolet, fill=cabgviolet, text width=36mm},
  hum/.style={box, draw=cared, fill=cabgred, line width=1.2pt, text width=36mm, minimum height=33mm},
  con/.style={box, draw=caindigo!70!black, fill=white, text width=36mm},
  cmp/.style={box, draw=caviolet, fill=cabgviolet!60, densely dashed, text width=34mm, inner sep=4.5pt},
  ver/.style={draw=caslate, fill=white, diamond, aspect=1.45, align=center, inner sep=2.5pt, minimum width=25mm, line width=1.2pt, drop shadow},
  badge/.style={circle, text=white, inner sep=1.5pt, font=\bfseries\footnotesize},
  flow/.style={-{Stealth[length=2.9mm]}, line width=1.3pt, draw=caslate},
  fate/.style={-{Stealth[length=3.2mm]}, line width=1.7pt, rounded corners=8pt},
  lbl/.style={fill=white, align=center, inner sep=2.6pt, font=\footnotesize\sffamily, rounded corners=2pt, draw=caslate!30, line width=0.5pt},
  fatelbl/.style={lbl, line width=0.8pt, inner sep=3.2pt}
]
\node[gen] (gen) at (2.05,-1.50) {\small\textbf{Generator}\\[-1pt] \small\textbf{research AI, vendor $A$}\\[2.5pt] e.g.\ Claude-based. Plans and runs the experiment; writes one \emph{increment}: data, metadata, logs, provenance.};
\node[sci] (sci) at (8.10,-1.50) {\small\textbf{Science Repository}\\[2.5pt] The increment lands as commit \texttt{S}: frozen, \mbox{diffable}, citable. Every later step points at exactly \texttt{S}.};
\node[det] (dcl) at (14.15,-1.50) {\small\textbf{Script checks: no AI}\\[2.5pt] Plain code checks \texttt{S} first: schema? units? sources? convergence? A hard fail is final; no model can override it.};
\node[cmp] (cmp) at (8.10,-4.05) {Cloud compute keeps running: a verdict gates the \emph{next} increment, never live jobs.};
\node[hum] (hum) at (4.90,-6.75) {\small\textbf{Human principal}\\[2.5pt] The only person in the loop: writes the rulebook, rules on escalations, reads advisories asynchronously. Never interrupted otherwise.};
\node[con] (con) at (9.85,-6.75) {\small\textbf{The rulebook (Constitution)}\\[2.5pt] Human-authored, versioned. Every rule: stable ID $+$ decidable criterion. Changes land only \emph{between} cycles.};
\node[aud] (aur) at (14.35,-5.75) {\small\textbf{Auditor}\\[-1pt] \small\textbf{reviewing AI, vendor $B \neq A$}\\[2.5pt] e.g.\ GPT-based; shares no memory with $A$. Reads \texttt{S} rule by rule; every finding must cite a rule ID and evidence. BLOCKER findings gate; ADVISORY never do.};
\node[ver] (ver) at (9.50,-10.20) {\small\textbf{Verdict}};
\node[led] (led) at (14.35,-10.20) {\small\textbf{Audit ledger}\\[2.5pt] \texttt{report.md} $+$ \texttt{receipt.json}, append-only. The receipt pins commit \texttt{S}, file hashes, rulebook version, auditor identity.};
\node[badge, fill=cagreen]  at (gen.north west) {1};
\node[badge, fill=caindigo] at (sci.north west) {2};
\node[badge, fill=caslate]  at (dcl.north west) {3};
\node[badge, fill=caorange] at (aur.north west) {4};
\node[badge, fill=caviolet] at (led.north west) {5};
\node[badge, fill=caslate, anchor=south west] at ([xshift=-0.5mm, yshift=0.2mm]ver.north east) {6};
\node[badge, fill=cared]    at (hum.north west) {7};
\draw[flow] (gen) -- node[lbl]{pushes} (sci);
\draw[flow] (sci) -- node[lbl]{audit trigger,\\[-2pt] pinned to\\[-2pt] exactly \texttt{S}} (dcl);
\draw[flow] (dcl.south -| aur.north) -- node[lbl, inner sep=2pt, yshift=0.8mm]{all scripts pass} (aur.north);
\draw[flow] (aur.south) -- node[lbl, inner sep=2pt]{commits report $+$ receipt} (led.north);
\draw[flow] (led.west) -- node[lbl, yshift=4.8mm, xshift=-1.5mm]{verdict\\[-2pt] sent back} (ver.east);
\draw[flow, densely dashed, draw=caviolet] (sci) -- (cmp);
\draw[flow, densely dashed, draw=cared!70!black] (hum) -- (con);
\draw[flow, draw=caindigo!70!black] (con.east) -- (con.east -| aur.west);
\draw[fate, draw=cagreen!75!black] (ver.south) -- (9.50,-11.55) -- (1.70,-11.55) -- (1.70,-3.02);
\node[fatelbl, draw=cagreen!60] at (5.60,-11.55) {\textbf{\textcolor{cagreen!60!black}{PASS}}: receipt re-verified at the admission gate\\[-1pt] $\Rightarrow$ \texttt{S} admitted; the next increment builds on it};
\draw[fate, draw=caorange!85!black] (ver.west) -- (2.55,-10.20) -- (2.55,-3.02);
\node[fatelbl, draw=caorange!60] at (5.45,-10.20) {\textbf{\textcolor{caorange!80!black}{BLOCKED}}: revise or dispute\\[-1pt] (once per finding); round ${+}1$, at most 3};
\draw[fate, draw=cared] (ver.north) -- (9.50,-9.05) -- (5.50,-9.05) -- (5.50,-8.42);
\node[fatelbl, draw=cared!55] at (6.00,-9.12) {\textbf{\textcolor{cared}{ESCALATE}}: rounds spent, invalid\\[-1pt] audit, or either agent asks};
\node[align=center, font=\footnotesize\sffamily, text=caslate, anchor=north] at (8.1,-12.15)
  {Every element on this page is a git commit or a GitHub issue: the parsed supervision record is re-inspectable by anyone the operator grants access to.};
\end{tikzpicture}}
\caption{One experiment increment through the loop, read clockwise from the top left. Numbers give the order, and the three coloured paths out of the verdict are the only ways a cycle can end. The report commit lands before its receipt: the receipt binds the report's content hash, so it can never reference an uncommitted object, and the receipt is derived by the deterministic audit-side workflow, never authored by the model. Who may write where is Figure~\ref{fig:powers}; the walk in full is \S\ref{sec:roles} and \S\ref{sec:rounds}.}
\label{fig:arch}
\end{figure}

The two-repository split is separation of powers made mechanical (Figure~\ref{fig:powers}): the Auditor holds no write access to the Science Repository, the Generator none to the Audit Repository, so neither agent holds credentials to alter the other's record. Guarding each record against its \emph{own} author also requires branch protection.

\subsection{Eight invariants}\label{sec:invariants}

An implementation conforms to the extent that it preserves the following; Table~\ref{tab:invariants} grades, invariant by invariant, what the reference implementation enforces, targets, or leaves to deployment, and we reserve ``invariant'' for the protocol's obligation, never for an implementation's achievement. (\invariant{7} and \invariant{8} are revision-1 invariants, adopted from the cross-vendor audit of this repository described in \S\ref{sec:pilot}.)

\begin{description}[itemsep=2pt,topsep=3pt]
  \item[\invariant{1} --- Heterogeneity.] The Auditor's model family differs from the Generator's. This prevents the auditor from ever judging its own generations, the configuration in which self-preference bias has been demonstrated \cite{panickssery2024}. Similarity-driven preference is reduced, not eliminated. It is likewise designed to decorrelate vendor-idiosyncratic failure modes, an effect this paper does not measure (\S\ref{sec:implementation}). Heterogeneity is also one of three isolations the protocol stacks: \emph{parametric} (different weights), \emph{contextual} (no shared memory: the Auditor never sees the Generator's prompts, reasoning, or conversational state, only committed artefacts). The mechanism we intend to remove is the one by which shared context sways a judge. The nearest evidence, that assistants accommodate a user's stated view \cite{sharma2023sycophancy}, concerns human-prompted sycophancy, not agent-to-agent anchoring, so contextual isolation is a design precaution here and not a channel we have shown to be closed, and \emph{credential} (no shared credentials; \S\ref{sec:roles}).
  \item[\invariant{2} --- Ledger completeness.] Every protocol artefact (increment, report, verdict, dispute, escalation, and every change to the Constitution) is a commit or an issue in one of the two repositories. The guarantee has three tiers and only the first is currently met. \emph{Parsed-record inspectability}: the reports, verdicts and receipts as committed, which the reference implementation does provide. \emph{Raw-exchange preservation}: the model requests and responses behind them, which it does not yet capture. \emph{Full process replay}: re-executing the supervision from the record, which is not achievable at all while the model calls are stochastic. A deployment should say which tier it holds. Ours holds the first, and \S\ref{sec:implementation} records two further gaps: the deployed variant's controller state lives outside both repositories, and its generator's session state was never committed.
  \item[\invariant{3} --- Citation validity.] An audit report must cite the rule IDs it applied and the commit hash of the Constitution version in force. A report that cites nothing, or an unverifiable version, is \emph{invalid}, is treated as an Auditor failure, and escalates. It can never pass an increment. The precedence is part of the invariant: an invalid report escalates even where the deterministic layer has independently blocked, because an integrity failure must never be absorbed into an ordinary scientific verdict, and admission fails closed on it.
  \item[\invariant{4} --- Determinism first.] The DCL runs before any LLM audit. A DCL hard failure yields \texttt{BLOCKED} regardless of any model's opinion, and no model may waive a DCL verdict. Rationale: heterogeneous models still train on largely overlapping public corpora and hence can share their errors. The supervision layer whose failure modes are least entangled with any model is one with no learned component at all.
  \item[\invariant{5} --- Bounded revision.] At most \texttt{max\_rounds} Generator--Auditor exchanges per increment (default three). Exhaustion escalates. This is the oscillation guard: without it, two mutually-correcting agents can disagree indefinitely, each round locally plausible and the pair jointly divergent.
  \item[\invariant{6} --- Graded interruption.] Findings carry exactly two severities. \texttt{BLOCKER} (objective defect: DCL failure, internal contradiction, missing provenance, method--declaration mismatch) gates the increment. \texttt{ADVISORY} (judgement: parameter taste, style, scope) is recorded for asynchronous human reading and never gates. Humans are interrupted only by \texttt{ESCALATE}. Running compute is never interrupted at all.
  \item[\invariant{7} --- Receipt binding.] Every verdict is a \emph{receipt} bound to the audited commit SHA; to an artefact manifest (paths and content hashes) that the Audit Repository derives from that SHA itself, never from caller-supplied payload; to the Constitution, check-layer, and prompt versions applied; to the auditor's declared identity; and to the ledger commit of the report. A verdict that binds less is not a receipt.
  \item[\invariant{8} --- Fail-closed admission.] Protected actions (admitting the next increment, production submission, claim publication) proceed only on verification of a valid, current, matching receipt. Absent, stale, conflicting, unbound, or model-audit-less receipts deny by default and escalate: no audit, no admission.
\end{description}

Two severities only. In our judgement every additional level invites severity-inflation negotiation between the agents. Where a finer distinction is needed it belongs in the rule's acceptance criterion, not in the escalation lattice.

Table~\ref{tab:invariants} condenses the eight invariants into a reference card: the guarantee each one makes, the mechanism that enforces it, and how far the reference implementation carries that enforcement today. The statuses are the ones \S\ref{sec:implementation} accounts for. The distance that remains is enumerated in \texttt{ROADMAP-R2.md}.

\begin{table}[t]
\centering
\footnotesize
\caption{The eight invariants as a reference card. ``Enforced'' means the reference implementation refuses the violating action in code; statuses follow the audited account in \S\ref{sec:implementation}, and the implementation as a whole targets \invariant{1}--\invariant{8} while implementing a subset.}
\label{tab:invariants}
\begin{tabular}{@{}l p{3.45cm} p{4.35cm} p{4.35cm}@{}}
\toprule
 & Guarantee & Enforced by & Implementation status \\
\midrule
\invariant{1} & The Auditor never judges its own model family's output & Vendor pairing asserted from \texttt{crossaudit.yml} at dispatch & Enforced from configuration; vendor identity self-declared (\S\ref{sec:threat}) \\
\invariant{2} & Supervision history replayable from the two repositories alone & Every artefact a commit or issue; append-only full-SHA cycle directories & Parsed reports preserved; raw model exchanges not yet captured \\
\invariant{3} & A report citing no rules, or an unverifiable Constitution, can never pass & Invalid replies escalate, alone or atop a DCL hard failure (verdict-precedence tests); admission refuses receipts whose \texttt{audit\_integrity} is not \texttt{OK} (tested) & Enforced \\
\invariant{4} & No model may waive a scripted hard failure & Verdict synthesis is code; a DCL hard failure dominates any model verdict & Enforced \\
\invariant{5} & Generator--Auditor oscillation terminates & Controller-managed rounds; exhaustion escalates & Enforced by the controller state machine \\
\invariant{6} & Humans are interrupted only by escalation & Two-severity schema; \texttt{ADVISORY} never gates & Enforced in schema and router \\
\invariant{7} & A verdict binds commit, manifest, and versions, or it is not a receipt & Receipt derived from the checked-out tree; verifier checks every bound field & Receipts emitted and verified; raw-exchange capture short of \invariant{2}'s full intent \\
\invariant{8} & No audit, no admission & Single-use, freshness-guarded admission; \texttt{DCL\_ONLY} never admits & Verifier and controller enforce; binding admission to branch protection is a deployer toggle \\
\bottomrule
\end{tabular}
\end{table}

\subsection{The Constitution}\label{sec:constitution}

The Constitution is the protocol's quality ceiling: the Auditor's blocking power may not, by rule, exceed the rules it is given (miscitation is caught by the dispute channel, not prevented), and the Generator is free to exploit whatever the rules fail to say. CrossAudit therefore treats it as law rather than documentation. Every rule carries a stable identifier (\rulename{CA-DATA-001}), a severity, a path scope, a \emph{decidable} acceptance criterion, and a rationale. Decidability is the load-bearing requirement: a criterion a competent reader can evaluate against named artefacts without further judgement. What cannot be phrased decidably is either demoted to \texttt{ADVISORY} or (better) mechanised and moved into the DCL. Meta-rules govern the audit itself: missing evidence is a finding, never an implicit pass (\rulename{CA-META-001}); a report without citations is invalid (\rulename{CA-META-002}, enforcing \invariant{3}); the Auditor may not issue a \texttt{BLOCKER} without citing a \texttt{BLOCKER}-severity rule and so does not legislate taste (\rulename{CA-META-003}); and matters beyond the increment-plus-Constitution evidence base must escalate rather than be guessed (\rulename{CA-META-004}). Constitution changes are commits reviewed by the human principal. The loop can propose rules but never enact them. (The name deliberately echoes, but must not be confused with, Constitutional AI \cite{bai2022constitutional}: there, a constitution steers a model's training-time self-critique. Here it is a runtime rulebook applied by a structurally independent auditor and versioned in the ledger.)

\subsection{Rounds, disputes, termination}\label{sec:rounds}

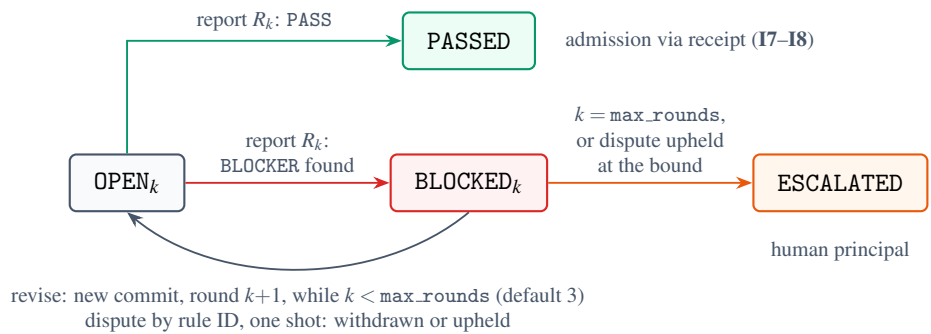
\begin{figure}[t]
\centering
\begin{tikzpicture}[
  font=\small, >=Stealth, line width=0.7pt,
  st/.style={draw, rounded corners=2.5pt, minimum height=7.5mm, inner xsep=3.2mm, thick},
  lab/.style={font=\scriptsize, text=caslate, align=center},
  edge/.style={->, draw=caslate}]
  \node[st, draw=caslate, fill=cabgslate] (open) {\texttt{OPEN}$_{k}$};
  \node[st, draw=cared, fill=cabgred, right=27mm of open] (blocked) {\texttt{BLOCKED}$_{k}$};
  \node[st, draw=cagreen, fill=cabggreen, above=11mm of blocked] (passed) {\texttt{PASSED}};
  \node[st, draw=caorange, fill=cabgorange, right=27mm of blocked] (esc) {\texttt{ESCALATED}};
  \draw[edge, draw=cagreen] (open) |- node[lab, pos=0.75, above] {report $R_k$: \texttt{PASS}} (passed);
  \draw[edge, draw=cared] (open) -- node[lab, above] {report $R_k$:\\\texttt{BLOCKER} found} (blocked);
  \draw[edge] (blocked.south) .. controls +(-12mm,-10mm) and +(12mm,-10mm) .. node[lab, below=1mm] {revise: new commit, round $k{+}1$, while $k < \texttt{max\_rounds}$ (default 3)\\dispute by rule ID, one shot: withdrawn or upheld} (open.south);
  \draw[edge, draw=caorange] (blocked) -- node[lab, above] {$k = \texttt{max\_rounds}$,\\or dispute upheld\\at the bound} (esc);
  \node[lab, right=2.5mm of passed] {admission via receipt (\invariant{7}--\invariant{8})};
  \node[lab, below=2.5mm of esc] {human principal};
\end{tikzpicture}
\caption{Termination as a walk on four states. Every path from \texttt{OPEN} ends in \texttt{PASSED} (admission gated by receipt) or \texttt{ESCALATED} (a human decides); the only cycle in the graph is the revision loop, and the round bound cuts it after \texttt{max\_rounds} exchanges. The dispute edge re-enters the same loop instead of adding one: a finding may return to the Auditor once.}
\label{fig:termination}
\end{figure}

Round $k$ for an increment: the Auditor publishes report $R_k$ with a verdict. On \texttt{BLOCKED}, the Generator either \emph{revises} (a new commit, opening round $k{+}1$) or a dispute is lodged against specific findings by rule ID, with grounds supplied by the principal. A disputed finding returns to the Auditor once: it is withdrawn with a logged reason, or upheld. An upheld dispute at the round bound escalates. Disputes exist because the Auditor is fallible in both directions: a false-positive \texttt{BLOCKER}, silently obeyed, degrades the science as surely as a false negative. The one-shot rule prevents the dispute channel from becoming the oscillation it exists to control. Two asymmetries make the channel safe, and the product line (\S\ref{sec:productline}) enforces both in code, not by convention. The disputer states grounds and the Auditor rules, so nothing in a dispute can overturn a finding by itself. It buys a second reading, and the second reading is the Auditor's. And the Generator may not dispute at all. Grounds come from the principal, because a generator arguing with its own audit is the anchoring channel \invariant{1}'s contextual isolation exists to close. Contesting the \emph{rule} rather than the finding is an amendment, which is dated and effective only between cycles. Termination is structural: each increment ends in \texttt{PASS} or in human hands after at most \texttt{max\_rounds} exchanges. Figure~\ref{fig:termination} draws the walk.

\begin{table}[t]
\centering\footnotesize\setlength{\tabcolsep}{4pt}
\caption{Four objects, and which claims attach to which. The paper moves between them, and a claim about one is not a claim about another. Only the trial in \S\ref{sec:pilot} was measured, and it measured the reference scorer.}
\label{tab:objects}
\begin{tabular}{@{}p{22mm}p{34mm}p{34mm}p{34mm}p{22mm}@{}}
\toprule
Object & What it is & What it is claimed to do & What it is not & Evidence here \\
\midrule
Protocol & The eight invariants, \S\ref{sec:invariants} & Define a supervision configuration & An implementation & Argument only \\
Reference implementation & \S\ref{sec:implementation}, workflows and Python & Target all eight, enforce a subset (Table~\ref{tab:invariants}) & A product & Regression tests; the trial's scorer \\
Live deployment & \S\ref{sec:implementation}, one chemistry pipeline & Show the loop runs unattended & Conforming: it diverges on \invariant{2} and on two-level severity, both named & Observational ledger, no control \\
Product line & \S\ref{sec:productline}, a separate package & Show what a second implementation taught the specification & Evaluated anywhere in this paper & None \\
\bottomrule
\end{tabular}
\end{table}

\section{Reference implementation and deployment}\label{sec:implementation}

\subsection{Mechanics}

The reference implementation\footnote{\url{https://github.com/dongzhaohe321418-lab/crossaudit} (MIT licence).} uses no infrastructure beyond GitHub itself and two model API endpoints. A push under \texttt{experiments/} triggers a cross-repository dispatch carrying the commit SHA; the audit side derives the audited scope from the SHA itself and the round from its own report index; caller-supplied payload is treated as an untrusted hint. The Audit Repository's workflow checks out the science repository \emph{pinned at the audited SHA}, asserts \invariant{1} from configuration, runs the DCL, then invokes the Auditor via an OpenAI-compatible endpoint with the Constitution, the DCL output, and the increment content as fenced data. The report is committed to the ledger, the verdict is dispatched back, and the science repository routes it, proceed, revise, or open an escalation issue. The whole implementation is three workflow files and two Python programs (a check runner and an auditor runner) of a few hundred dependency-light lines. Configuration is a single \texttt{crossaudit.yml} declaring vendors, \texttt{max\_rounds}, and escalation policy.

\paragraph{Status against the invariants.} Per-invariant enforcement and status are Table~\ref{tab:invariants}'s job (\S\ref{sec:invariants}); what belongs here is how those statuses were earned, because the route is itself evidence for the protocol. Following two rounds of cross-vendor audit of this repository (\S\ref{sec:pilot}; both reports and dispositions committed under \texttt{audits/}), scope became SHA-derived, rounds ledger-derived rather than generator-controlled, receipts full-field, and the reply validator strict; truncated, unreadable, or symlinked inputs now escalate rather than pass, and keyless runs yield \texttt{DCL\_ONLY}, which the router treats as non-admission. A third cross-vendor read (also committed) then found the sharper failure mode: the controller state machine and full receipt verifier existed but were not wired into the execution path, dry-run admission, a hard-coded report path that broke on disputed rounds, a constitution check that could never fire. Those wiring defects are fixed. What remains unimplemented is the deployer-side binding of the admission status to branch protection, and raw model-exchange capture in the audit runner. The public implementation should therefore be read as \emph{targeting} \invariant{1}--\invariant{8} and implementing selected checks. The remaining distance is enumerated in \texttt{ROADMAP-R2.md}. \paragraph{Three details with protocol weight.} First, the auditor runner validates the model's reply against a fixed schema (verdict, findings with severities and rule citations) and any malformed or citation-free reply is converted to \texttt{ESCALATE} (\invariant{3}); a familiar failure of LLM judges (the confident empty report) is thereby converted from silent passage into an escalation. Where the DCL has also hard-failed, the recorded verdict now escalates as well: the synthesis places \invariant{3} above \invariant{4} deliberately, because an integrity failure must never be absorbed into an ordinary scientific verdict, and a regression test pins the combined case (\texttt{controller/tests/test\_verdict\_precedence.py}; a valid model \texttt{PASS} still cannot outrank a DCL failure, so \invariant{4} is untouched). The receipt carries the auditor failure in \texttt{audit\_integrity}, and admission refuses any receipt where it is not \texttt{OK} (also tested), so the gate holds even against an older deployed runner. Second, increment content is passed to the Auditor as fenced, size-bounded content with the system prompt instructing that embedded instructions be ignored, a best-effort defence whose residual risk \S\ref{sec:threat} states plainly. The reference Auditor has no tools. Third, verdict synthesis is code, not model output: DCL hard failures dominate any model verdict (\invariant{4}).

\subsection{A live deployment}

A closely related variant of CrossAudit runs live in the first author's computational-chemistry pipeline (perovskite additive screening). The generation side is a Claude-based research agent that plans calculations, launches them on Alibaba Cloud E-HPC and a GPU worker, and pushes every increment to a private science repository whose audit policy, an epistemic-status ledger separating established, bounded, and provisional claims, together with a standing retraction log, is itself a committed, versioned document. A controller opens each cycle against a fixed science commit. The audit request pins a SHA-256 manifest of every tracked file. A Codex-CLI auditor (OpenAI lineage, so \invariant{1} holds against the Claude-based generator) writes four artefacts into an append-only per-cycle directory in the audit repository: a human-readable report, a machine-readable result, run metadata, and a per-file hash manifest. All sit on one long-lived \texttt{audit} branch that later cycles may not rewrite. The controller enforces this by validating every audit commit before recording a cycle as final, which detects a rewrite rather than preventing one. Prevention would need branch protection. The controller's own orchestration state deliberately lives outside both repositories, so tampering by either agent with the record of its own supervision is detectable against controller-held state. 

Three deliberate divergences from the reference design should be named plainly: the deployed auditor is a tool-bearing CLI agent rather than the reference's toolless one, contained instead by that commit validation and the append-only cycle layout. The externalised controller state means the deployment's history is fully replayable from the repositories \emph{plus} that state, a weaker property than \invariant{2}'s repositories-alone ideal; and findings carry a four-level severity ladder feeding scope-level gates, rather than the protocol's two levels, a live experiment running against \S\ref{sec:invariants}'s two-level argument, whose outcome the telemetry will eventually adjudicate. 

Every committed scientific artefact and parsed supervision decision lives on GitHub; raw model exchanges, transient progress and controller orchestration state do not (\invariant{2}). The system is designed so that any single component (generator, auditor, controller host, workstation) can fail without loss of the committed ledger state.

Two refinements of the deployed system go beyond the reference implementation and are worth adopting upstream. First, findings are \emph{evidence-typed}: each carries a policy citation (section and rule ID), a deterministic check identifier where applicable, and a copy-paste \emph{reproduce} command, so a human can re-derive any deterministic finding in seconds. Second, blocking is \emph{scope-granular}: a finding blocks named scopes, production-job submission, claim publication, rather than the pipeline wholesale, refining \invariant{6}'s graded-interruption principle. Human authority enters through escalation, plus one deployment-specific addition: a standing principal-approval gate on production launches, recorded in the ledger. The loop is live in the plainest sense: its first cycle (opened and closed within six minutes) returned a \texttt{BLOCK} decision on deterministic grounds (manifest-hash disagreements and commit-provenance defects), freezing production submission and claim publication until resolved. Audits gate the record and future submissions, never running jobs: E-HPC work in flight is never interrupted. A redacted public mirror of the deployment's finalised cycles is maintained in the reference repository (\texttt{ledger-mirror/}; scientific magnitudes masked, structure, rule citations and reproduce-command shapes verbatim). Including that first cycle, the ledger accumulated seven model-audited cycles across two days (Figure~\ref{fig:trace}): decisions ran \texttt{BLOCK} five times, then \texttt{PASS WITH CAVEATS}, then \texttt{PASS}, while findings per cycle fell 7, 4, 3, 2, 1, 1, 0. Part of that fall is the audit boundary moving, because the early cycles audit accumulated backlog and the later ones fresh increments; and with no control condition, what would have been caught without the loop is not knowable from this series. Of fourteen distinct findings, twelve are behaviourally confirmed real: each revise leg is a science-repository commit naming the finding IDs it closes, and the following cycle verifies the closure; two remain unresolved at the data freeze. All twelve were detected and closed before the gated downstream action. Whether they would have been caught some other way, this deployment cannot say: it ran no control condition. One leg (the ledger's numbering gap at cycle~3) ran as a deterministic-tier follow-up without a model audit and is reported as-is. The generator's session state was never committed, so no in-context self-review arm can be reconstructed retrospectively; that absence is itself evidence for \invariant{2}. The mined, structurally redacted cycle records are frozen in this repository (\texttt{experiment/v3/real-ledger/}).

The loop has since kept running. By 3~August (second freeze: science \texttt{485106fa}, audit \texttt{abb10a47}; the counting rules and regeneration script are committed with it) the ledger held 21 model-audited cycles carrying 39 finding occurrences under 23 distinct finding identifiers, 20 of the 23 behaviourally confirmed closed (a fix commit named the finding and a re-audit verified the closure, which is closure evidence, not independent ground truth) and 3 still open at the freeze. The first arc converged, then two more increments passed on their first audit. What followed matters more. Sustained development settled at roughly one to three findings per cycle, and stayed there. The loop reaches a working rate, not zero, and that is the regime a long-running deployment lives in. Median closure lag, counted per distinct finding from first raise to verified closure over the 20 closed (the 3 open are excluded, not imputed), was one cycle, with a mean of 2.10 and a maximum of 7; ten of the 23 identifiers recur across cycles, which is what a re-audit exists to catch. On the deployment's own four-level ladder the 23 findings graded as three \texttt{CRITICAL}, thirteen \texttt{HIGH}, four \texttt{MEDIUM}, and three \texttt{LOW}. Twenty-five of the 39 occurrences cite a single rule, firing in 16 of the 21 cycles, the kind of concentration that tells a principal where the next deterministic check should go. The scope-granular gate was working throughout: counting finding--scope incidences, which overlap because one finding can gate two scopes, 33 gated claim publication and 12 gated production submission, and at the snapshot a claim-publication action stood withheld pending a clean verdict beneath it. Audits took a median of six minutes (range 3.7--18.2); the interval from one audit finishing to the next beginning, which bundles the repair with any idle time, had a median of thirty-four minutes (quartiles 18 and 70, the maximum an overnight gap). So the audit is not the wall-clock bottleneck. What those repairs cost in human attention, which is what \S\ref{sec:discussion}'s economics is actually about, the ledger does not record.

Four things temper all of this. We report the shape of that second stretch and claim nothing causal from it, because the two stretches are not a matched comparison: the first audited accumulated work, the second audits fresh increments, and we drew the boundary after seeing the series. The auditor model was not held fixed either, since four identifiers appear in the metadata, so the series mixes configurations and cannot support arm comparison. Findings carry the deployment's own source label, 11 of the 39 occurrences marked \texttt{DETERMINISTIC} against 28 marked \texttt{JUDGMENT}; that records what caught each defect, not what could have. The second question is the one our registered study asks, and it remains unanswered. Finally, the v3 study's Part~B labels stay frozen at the 31~July snapshot. These later cycles are observational telemetry from one deployment with no control condition, and the analysis behind them, limits stated at length, is committed at \texttt{experiment/v3/real-ledger-b/}.

Costs for the reference design are modest. The DCL is free. A single fenced-increment audit call scales with the size of the increment, not of the computation behind it, a few hundred lines of results and metadata read per call, while the experiment it gates may represent hours of HPC time; we quote no currency figure, because per-cycle token telemetry is deferred and an unmeasured price is not a datum. (The deployed variant's tool-bearing auditor reads more and costs more; per-cycle cost reporting is deferred with the rest of the telemetry.) Auditing costs roughly what it costs to read the documentation, not what the compute behind it cost. Two caveats: multi-round exchanges multiply audit calls, and escalations spend human time, which is the truly scarce resource here.

The deployment's role is existence proof, an observational feasibility case study with no control condition and no effectiveness claim, for the deployed variant: the loop has run unattended, at single-researcher scale, on commodity infrastructure, and has completed one full convergence arc from \texttt{BLOCK} to \texttt{PASS}. Its first catch was attributable to the deterministic layer; the thirteen findings that followed came through the tool-bearing auditor, and what fraction of them a script could have caught (the mechanisability of the confirmed set) is a registered endpoint of the follow-up study, adjudication pending, not a claim made here.

\begin{figure}[t]
\centering
\resizebox{\textwidth}{!}{%
\begin{tikzpicture}[x=1cm, y=1cm,
  font=\footnotesize\sffamily,
  chip/.style={draw=caindigo, fill=cabgindigo, rounded corners=2.5pt, inner sep=4pt, font=\scriptsize\ttfamily},
  vbox/.style={draw, rounded corners=2.5pt, align=center, inner sep=4pt, font=\scriptsize\sffamily, minimum width=17mm, line width=0.8pt},
  vb/.style={vbox, draw=cared, fill=cabgred},
  vc/.style={vbox, draw=caorange, fill=cabgorange},
  vp/.style={vbox, draw=cagreen, fill=cabggreen},
  cyc/.style={circle, fill=caslate, text=white, inner sep=1.5pt, font=\bfseries\scriptsize},
  down/.style={-{Stealth[length=2mm]}, line width=0.9pt, draw=caslate},
  nxt/.style={-{Stealth[length=2.2mm]}, line width=1pt, draw=caslate!85}
]
\draw[densely dashed, caslate!45] (1.275,0.42) -- (1.275,-3.55);
\node[font=\scriptsize\sffamily, text=caslate!80, anchor=south east] at (1.15,0.46) {30 Jul};
\node[font=\scriptsize\sffamily, text=caslate!80, anchor=south west] at (1.40,0.46) {31 Jul};
\foreach \i/\sha/\style/\cyc/\lab/\n/\bc in {
  0/81fd505e/vb/1/{BLOCK\\7 findings}/7/cared,
  1/bfece650/vb/2/{BLOCK\\4 findings}/4/cared,
  2/746564b6/vb/4/{BLOCK\\3 findings}/3/cared,
  3/b7de4784/vb/5/{BLOCK\\2 findings}/2/cared,
  4/4aed10a9/vb/6/{BLOCK\\1 finding}/1/cared,
  5/4ccb0cd6/vc/7/{PASS w/\,caveats\\1 finding}/1/caorange,
  6/af0dc587/vp/8/{PASS\\0 findings}/0/cagreen}{
  \node[chip] (s\i) at (\i*2.55,0) {\sha};
  \node[\style] (v\i) at (\i*2.55,-1.45) {\lab};
  \node[cyc, anchor=south west] at ([xshift=-1mm, yshift=-0.5mm]v\i.north west) {\cyc};
  \draw[down] (s\i) -- (v\i);
  \ifnum\n>0
    \fill[\bc!75] ({\i*2.55-0.21},-3.55) rectangle ({\i*2.55+0.21},{-3.55+\n*0.155});
  \fi
  \node[font=\bfseries\scriptsize\sffamily, text=\bc!85!black, anchor=south] at ({\i*2.55},{-3.55+\n*0.155}) {\n};
}
\draw[caslate!35, line width=0.6pt] (-0.85,-3.55) -- (16.15,-3.55);
\node[font=\scriptsize\sffamily, text=caslate, anchor=east, align=right] at (-0.50,-2.90) {findings\\per cycle};
\foreach \i [evaluate=\i as \j using int(\i+1)] in {0,...,5}{
  \draw[nxt] (v\i.east) -- node[font=\scriptsize\sffamily, text=caslate, fill=white, inner sep=1pt]{fixes} ([yshift=0mm]s\j.west);
}
\node[font=\scriptsize\sffamily, text=caslate, anchor=south] (tiernote) at (3.825,0.95) {cycle 3: Tier-0 follow-up, no model audit};
\draw[densely dashed, caslate!60, -{Stealth[length=1.6mm]}] (tiernote.south) -- (3.825,0.30);
\node[font=\scriptsize\sffamily, text=caslate, anchor=north] at (7.65,-3.95)
  {Each next science commit names the finding IDs it closes; the following cycle verifies the closure. 12 of 14 distinct findings confirmed closed; 2 unresolved at freeze.};
\end{tikzpicture}}
\caption{The live deployment's first convergence arc (seven cycles; the 21-cycle figures in the text come from a second freeze, 2026-08-03), mined from its audit ledger at the first data freeze (science \texttt{af0dc587}, audit \texttt{88b92429}; structurally redacted records in \texttt{experiment/v3/real-ledger/}). Top lane: the audited science commits, with the day boundary from the ledger's timestamps dashed. Middle lane: the audit cycles, coloured by decision, with the deployment's own numbering gap preserved (the dashed pointer marks leg~3, a deterministic-tier follow-up without a model audit). Bottom: findings per cycle on a shared baseline, falling 7, 4, 3, 2, 1, 1 to 0 as decisions move from \texttt{BLOCK} to \texttt{PASS}; the twelve behaviourally confirmed defects were closed before the downstream actions they gated.}
\label{fig:trace}
\end{figure}

\subsection{What a second implementation taught the specification}\label{sec:productline}

A separate line of work rebuilt the loop as an installable package, carrying the engine's invariant semantics over unchanged.\footnote{\url{https://github.com/dongzhaohe321418-lab/crossaudit_v3} (MIT licence), version 3.2.0, at which this paper's evidence about the product line is frozen; later product versions are outside it. The research repository does not chase the product, and the protocol's authoritative text remains this paper.} It is measured nowhere here, and we describe none of its interface. What a second implementation buys is the ordinary benefit of writing a specification down twice: the underspecified parts of the first writing become visible. Three did, and all three are about the protocol, not the package.

\paragraph{Admission is a tier, not a switch.} \invariant{8} says protected actions proceed only on a verified receipt, and \S\ref{sec:invariants} notes that binding the check to branch protection is a deployer action. That framing is too binary for the field. Deployments sit at different distances from the guarantee: the history may be the operator's to rewrite, or beyond unilateral control, or privilege-separated between the agents, or genuinely gated so that a failed audit refuses the merge. The distance that matters most is the one between the last two, where a paired deployment publishes verdicts that refuse nothing. Calling that state \emph{paired} invites the reading \emph{gated}, and announcing a verdict while implying refusal is worse than saying nothing: it manufactures confidence the ledger cannot support. Naming it separately is the fix, and the evidence rule behind the naming matters as much: absent evidence never counts in favour, so persistence is established by taking the lock rather than by reading a configuration field, and branch rules are read from the platform rather than from the deployment plan that was meant to set them.

\paragraph{Whatever decides what the auditor sees is inside the trusted base.} A conversational front end needs something that decides whether an utterance changes the work, changes the rules, contests a finding, or asks a question. That component is a real addition to what must be trusted. One that could quietly decide what the auditor never sees would be a third agent, unaudited, holding the power the protocol exists to distribute. The resolution follows from \invariant{2} once the invariant is read as covering discretion as well as artefacts: the routing decision itself becomes a ledger entry, carrying the utterance, the choice, the confidence, and the action taken, so a misroute is visible and reversible instead of silent. The same reading forbids a subtler leak, since such a component must forward the principal's words and never the Generator's.

\paragraph{Domain generality is a claim an implementation can undermine quietly.} Nothing in \invariant{1}--\invariant{8} mentions science, but the reference deterministic layer's four checks assume experimental data, so the protocol's generality and its reference implementation's generality are not the same property. Separating them requires a domain-neutral default, domain packs as plugins, and one discipline that is easy to skip: where no pack exists for a field, the deterministic layer contributes almost nothing and the model audit carries the load alone. \invariant{4}'s mechanical guarantee thins, and an implementation in that position should say so instead of letting a uniform interface imply uniform strength. The same reticence governs a supervision display. A figure the ledger cannot support belongs on screen as \emph{not measured}, not as zero, because ``no escalations'' and ``we never looked'' are different claims and only one deserves a large font.

\subsection{An exploratory seeded-defect trial, audited and corrected}\label{sec:pilot}

We attempted a pre-registered, sealed seeded-defect pilot. The attempt did not survive contact with our own protocol. A cross-vendor audit of this repository, in which a second-vendor agent reviewed the generating agent's work, returned a BLOCK-grade report whose findings voided the ``sealed'' framing and several companion claims; the report and our dispositions are committed at \texttt{audits/}. We adopted its findings in full. What follows are the corrected, exploratory results, which proved more instructive than the experiment we had intended to run.

\paragraph{Design and execution.} A deterministic generator produced 30 synthetic increments in the reference format: 10 clean and 20 defective, carrying 43 seeded defects across ten registered classes, four of them mechanisable and six requiring reading. A SHA-256 digest of the defect key was committed at registration. Three arms then ran over identical inputs with the identical Constitution, though not under an identical harness, so vendor is confounded with harness and sampling throughout. The first was the deterministic check layer. The second ran fresh-context Claude subagents, one per increment, with no shared memory (session sampling, a deviation we disclose). The third ran \texttt{gpt-5.1} at temperature~0 under the frozen runner in CI. The corpus's author is the deterministic generator, so the arm labels below name the auditor configuration alone: no arm audits work that its own model family produced, and the own-generation setting in which self-preference is demonstrated \cite{panickssery2024} is not instantiated here.

\paragraph{What the audit found.} We checked each finding and accepted all of them. The key file was public from the registration commit onward. This was our mistake, a mis-pathed ignore rule, and the key is in any case derivable from the public seeded generator, so no blinding claim survives; what remains is procedural blinding of fresh-context auditors, which the ledger cannot verify. The headline Claude-arm score depended on a scoring-map expansion made after the outputs were committed. Under the frozen map the arm scores 38/43; under the adjudicated map, with corrections logged, it scores 43/43. The former is the reportable score and the latter is exploratory. The deterministic layer's five L6 catches contradict the registered taxonomy: the class was misregistered as model-only when its flag component is mechanisable. The trial is component-level throughout. The corpus was read directly, with no dispatch pipeline and no DCL precedence, so it is not an end-to-end run of the reference implementation. Finally, the trial's own outputs violate \invariant{3}: the runner recorded no Constitution hash, a defect one of the auditors itself flagged (below).

\begin{center}\small
\begin{tabular}{lcccc}
\toprule
Arm & \begin{tabular}{@{}c@{}}Recall\\lenient / strict\end{tabular} & \begin{tabular}{@{}c@{}}Chance floor\\lenient / strict\end{tabular} & \begin{tabular}{@{}c@{}}\textsc{blocker}\\findings\\on 10 clean\end{tabular} & \begin{tabular}{@{}c@{}}Verdict accuracy\\(of 30)\end{tabular} \\
\midrule
Deterministic layer & 17/43\,\tnotea & 4.8 / 4.2 & 0 & 25 \\
Claude-family (subagents) & 38 / 38\,\tnoteb & 22.4 / 16.3 & 0 & 30 \\
GPT-family (\texttt{gpt-5.1}, frozen runner) & 41 / 37\,\tnotec & 31.8 / 24.4 & 26 & 20 \\
\bottomrule
\end{tabular}

\vspace{2pt}
{\scriptsize All figures exploratory: blinding voided, key derivable from the public generator. \tnotea{} 12/12 of its registered scope $+$ 5/5 cross-hits on misregistered L6. \tnoteb{} 43/43 under post-hoc adjudicated scoring. \tnotec{} all ten clean increments blocked (10/10 false-block verdicts); many findings letter-valid on inspection; per-finding adjudication pending. Strict-tier scores assemble rule and location evidence across findings rather than binding one finding per defect, and so overstate defect-level localisation. Chance floors: 2000-shuffle permutation null, seed 20260731; \texttt{experiment/results/NULLCHECK.json}.}
\end{center}

\paragraph{The two arms read the rulebook differently.} The GPT-family arm scored 41/43 at the lenient tier but 37/43 at the strict tier. It emitted 113 findings to the Claude-family arm's 63, and it returned BLOCKED on \emph{all thirty} increments. That figure matters, because a stub auditor that blocks unconditionally would reproduce it exactly, including the 20/30 verdict accuracy. It is the reliability-without-validity failure in its purest form \cite{norman2026reliability}: a perfectly consistent judge carrying no information about the increment in front of it. Raw recall cannot settle the comparison, because it rises with the number of findings an arm emits; this is the same reason raw agreement overstates judge quality in the benchmarking literature \cite{norman2026reliability}. A permutation test (2000 shuffles of the increment-to-defect map, holding each arm's outputs fixed so that volume and citation habits are preserved; seed 20260731, \texttt{experiment/score\_nullcheck.py}) puts the random floors at 4.8, 22.4, and 31.8 of 43 for the three arms. The floors calibrate each arm against its own citation volume; they do not test the gap between arms, and an earlier draft said the lenient comparison was ``within noise'' with no statistic behind the phrase. The direct test, computed post hoc for this revision (\texttt{experiment/arm\_contrast.py}), pairs the two model arms on the 43 defects, which nest inside 20 defective increments; each reply is increment-level, so defects within an increment share one generation, and the primary inference therefore resamples and permutes increments rather than defects. At the lenient tier the arms disagree on seven defects (five caught only by the GPT-family configuration, two only by the Claude-family one); the recall difference is $+3$ of 43, cluster bootstrap 95\% CI $[-2, +8]$, cluster sign-flip permutation $p=0.46$ (the defect-level McNemar agrees at $p=0.45$ but overstates independence), so these data cannot distinguish the two configurations at that tier. At the strict tier the difference is $-1$ of 43 (cluster CI $[-6, +4]$, $p=1.0$; the table's footnote on blob-level matching applies) and the chance-corrected order reverses: floor-corrected agreement, $(\mathrm{observed}-\mathrm{floor})/(43-\mathrm{floor})$, is 0.81 for the Claude-family arm against 0.68 for the GPT-family arm. Inspection of the clean-increment blockers still suggests that many are letter-valid readings of an ambiguous rule, but per-finding adjudication was not logged, and the claim as originally framed was unfalsifiable: any disagreement would have counted as support. The rule at issue, \rulename{CA-DATA-001}, requires ``each numeric entry'' in the results file to carry a unit and a source; the corpus's convergence blocks carry units but no sources. The Claude-family auditor, like the check scripts and like the corpus author, read the rule by its intent and raised nothing. The foreign auditor enforced its letter. What survives is narrower but real, and it is about two configurations rather than two vendors. On identical inputs, one Claude-family and one GPT-family configuration read the same rulebook differently, which is consistent with \invariant{1}'s premise and does not establish it: a single model pair, run once, cannot separate a vendor effect from a prompt, temperature, or sampling effect. The trial nevertheless measures disagreement rather than discrimination, and a 10/10 false-block rate on clean increments means that in this configuration every increment escalates, so the $O(\text{escalations})$ economics of \S\ref{sec:discussion} hold only under a far better-calibrated auditor or a sharper Constitution. The practical outcome was a Constitution amendment: scope the rule decidably.

\paragraph{What we can still claim.} The limits should be stated plainly. This is a component-level, checklist-aligned trial (the prompt names the defect families the corpus injects) on template-generated data, with the Claude-family arm's execution self-attested and $n{=}43$. Zero blockers on ten clean increments bounds that arm's false-positive rate only below ${\approx}26\%$ (one-sided 95\%), and its advisory burden on the clean increments was five findings across five of ten. None of this is confirmatory evidence about detection rates. Three things the trial does establish. The model layers read: they caught 26 to 29 defects beyond the scripts' registered scope of twelve (21 to 24 beyond the seventeen the scripts caught in all). The two auditor configurations, drawn from different vendors, returned discordant readings of the same rulebook on identical inputs; with a deterministic corpus author, that is a fact about configurations, not about generator--auditor vendor pairings. And the repository's own disciplines (append-only outputs, logged adjudication, a committed cross-vendor BLOCK with itemised dispositions) were applied to the evaluation itself.

\paragraph{A note from the successor study's preparation.} The registered replacement for this trial has not run, but building its corpus produced one observation that bears on \invariant{4} directly, and we record it here rather than wait. A rule drafted for that study required the Auditor to apply any unit conversion and state its tolerance before alleging that two numbers disagree. It complied. On five clean increments it then quoted the correct conversion constant, read the correct inputs, and miscalculated the product by up to $1.8\times10^{-3}$\,eV while adjudicating a discrepancy of $8.8\times10^{-5}$\,eV, and the two increments it declared contradictory were the two where its own arithmetic was worst. A four-line script settles the same comparison correctly every time. One model, one vendor, five increments: this illustrates rather than establishes. What it illustrates is that the remedy for a model's arithmetic is not a rule demanding the model show more of it. A comparison that is decidable belongs to the deterministic layer, which is what \invariant{4} says; the rule was rewritten to route such pairs there instead, and the measurement is committed at \texttt{experiment/v3/SMOKE-FINDINGS.md}.

\subsection{What a reader can check}\label{sec:artefacts}

Four artefacts are versioned, and they answer different questions. Every statement below refers to the state of the reference repository at release tag \texttt{paper-v1.0.0}, cut at the source revision of this exact PDF, which is committed beside its source with its SHA-256 recorded in \texttt{paper/HASHES.txt}; ``exact'' is a checkable claim about the committed artefact and its hash, not about byte-level rebuild determinism, which TeX does not promise. An earlier draft pinned a bare commit instead, and the external review of this paper caught the draft having moved past its own pin; the anchor is now the tag that ships with the paper, so the two move together or not at all. The reference repository is public at that tag. The product-line repositories (\S\ref{sec:productline}) remain private at the time of writing, so claims about them rest on the audit excerpts committed under \texttt{audits/} and on the evidence grading of Table~\ref{tab:objects}, and a reader who cannot reach a cited artefact should discount the corresponding claim accordingly.

The reference implementation carries the protocol as specified here, with the controller, the receipt verifier, and their regression tests. The evaluation directory carries the trial corpus, the frozen scorer, and the permutation null with its seed, so every number in \S\ref{sec:pilot} regenerates from a clean checkout. The audit directory carries the cross-vendor reports written against this repository, and our dispositions on each one. That includes the findings that voided our own claims. The live deployment's supervision records are mined into the same repository in structurally redacted form: cycle identities, decisions, severities, rule citations, and finding lifecycles verbatim, scientific magnitudes masked.

\section{Threat model and limitations}\label{sec:threat}

Below is what CrossAudit withstands and what it does not.

\paragraph{Self-preference and vendor-correlated error.} These are addressed by \invariant{1}. \emph{Corpus}-correlated error is not: where the shared literature is wrong or outdated, both vendors can agree confidently and wrongly. The mitigation is \invariant{4} for everything mechanisable; for what is not, it is the human's asynchronous reading of \texttt{ADVISORY} trails. This residual cannot be assumed reducible by adding more LLM auditors, because all of them plausibly train on largely overlapping public corpora.

\paragraph{Prompt injection.} An increment could embed text addressing the Auditor. The Auditor runs toolless against a fixed reply schema, with increment content fenced as data; and because DCL verdicts are non-overridable, injection cannot waive the layer where hard defects are caught. Injection that shapes the \emph{LLM-side} report remains partially open, as it does for LLM-judge systems generally. This bullet describes the reference implementation; the deployed variant's tool-bearing auditor forgoes the toolless defence, a divergence \S\ref{sec:implementation} names together with its containment limits.

\paragraph{Auditor failure.} Empty, malformed, or rule-free reports convert to escalation (\invariant{3}), never to silent passage. The cost is human attention, which is the correct currency for an unreliable auditor.

\paragraph{Oscillation and deadlock.} These are bounded by \invariant{5}; the failure mode degrades to a human decision with the parsed exchange on the ledger, which is the intended floor.

\paragraph{Goodharting the Constitution.} A capable Generator will satisfy the letter of decidable rules. Decidability makes this visible rather than solved: the human reviews rule--outcome drift on the ledger and legislates accordingly. The Constitution is a living document, and its growth out of escalations is the system learning. A related boundary matters wherever a deployment lets the principal shape how the Generator works, through style guides, domain conventions, or worked examples. Such material may change how the work is done; it must never change where the Generator may write or who judges it. Guidance that could address the Auditor would be an unversioned rule, and guidance claiming wider write access than the path guard allows is only a text with an opinion.

\paragraph{Collusion.} Collusion has no out-of-band channel in the reference design. Coordination must ride the committed artefacts, where it is preserved for inspection. Encoded coordination \emph{within} artefacts, and any channels a tool-bearing auditor acquires, remain possible. The realistic soft risk is convergent style across many rounds, and it is met by the Constitution's review cadence.

\paragraph{The operator.} CrossAudit supervises the agents, not the principal. Selective publication, pre-publication history rewriting, and audit re-rolling by the operator all sit outside the protocol's guarantees. External anchoring (signed pushes, transparency logs) would extend tamper-evidence to the operator; until then, ``third-party-auditable'' should be read as \emph{internally consistent as published}. Relatedly, the DCL scripts and the Auditor's system prompt must live under the same ledgered, human-reviewed change control as the Constitution, because a Generator-editable check layer would hollow out \invariant{4}.

\paragraph{Vendor identity is self-declared.} The ledger proves that an exchange occurred; it does not prove the identity of the models behind OpenAI-compatible endpoints. \invariant{1} is asserted from configuration, not verified cryptographically. Attested inference or vendor-signed responses would close this gap; today it is an honest limit on what ``third-party-auditable'' can mean. (The reference example also shares one dispatch token across both workflows, so credential separation is likewise configuration, not proof.)

\paragraph{Confidentiality.} The protocol ships every increment, possibly unpublished and patentable material, to a second vendor's API. Data governance of that call (agreements, retention, disclosure risk) sits outside the protocol and must be settled per deployment.

\paragraph{Scope.} CrossAudit audits the research \emph{record}, not the world. It cannot detect a miscalibrated instrument, a wrong dataset honestly described, or fraud upstream of the repository. It is also currently specified for single-generator pipelines; multi-agent generation raises attribution questions we do not address here.

\section{Discussion}\label{sec:discussion}

\paragraph{Continuous integration for scientific claims.} The ambition is what CI achieved for software: not proof of correctness, but a cheap and inspectable floor under quality. Human judgement is then redeployed to where it discriminates (the rules, the escalations, and the science) and not spent uniformly on every increment. We believe the analogy holds in one further respect. CI spread without institutional permission; a single developer could adopt it unilaterally, and its artefacts, the badges and the logs, became social proof. One disanalogy must be owned: CI's checks are deterministic, while CrossAudit's distinctive layer is a stochastic judge. The DCL exists to keep the gate's hard core deterministic. CrossAudit is designed for the same adoption path: one researcher, two repositories and an afternoon for the audit loop itself, plus the CI, credential and branch-protection configuration that enforced admission needs.

\paragraph{Relation to platform-scale systems.} CrossAudit is complementary to, not competitive with, systems such as the AI co-scientist \cite{gottweis2025coscientist} or the AI Scientist \cite{lu2026nature}. Those systems generate. CrossAudit supervises whatever generates. The strongest version of our proposal is addressed to platforms of that kind. Their internal critique stages appear architecturally straightforward to re-instantiate across vendor lines and to externalise onto public ledgers, although the organisational and confidentiality costs are not ours to estimate. The resulting audit trails would give publishers and funders, who are currently asked to trust process descriptions \cite{naturenews2026}, something they can actually inspect.

\paragraph{What the ledger buys science socially.} A complete supervision history, public where the operator publishes it, changes the character of machine-generated results. A claim then arrives not as prose plus reputation but as prose accompanied by its audit record: which rules it was checked against, what was contested, and what a differently-trained model conceded or refused. Reviewers can spot-check the ledger instead of re-deriving trust from nothing.

\paragraph{The ledger as a data asset.} A CrossAudit deployment produces, as a by-product, something the field currently lacks: a corpus of \emph{supervised machine science}. Each tuple of increment, findings, dispute, and resolution is a labelled example of scientific work judged by an independent party under explicit rules, and therefore candidate training signal for better generator and auditor agents alike, with the attendant risk that training against the ledger is training against the rulebook, the optimisation \S\ref{sec:threat} names as Goodharting. Aggregated across deployments, failed audits would form a public error taxonomy of agentic research: what machine-generated science actually gets wrong, per rule, per field, over time. Whoever runs the loop accumulates this asset, and a community that runs it openly accumulates it as a commons.

\paragraph{Constitutions as community objects.} Because the Constitution is a plain, forkable file with stable rule IDs, it can outgrow any single deployment. A research community can maintain a shared domain constitution in the way communities maintain style guides and lint rules. A journal could publish the constitution it expects machine-assisted submissions to have been audited under, and accept the ledger link alongside the manuscript, much as data-availability statements are accepted today. Funders could write ``public supervision ledger'' into their terms. For regulated settings, pharmaceutical and clinical computation among them, the ledger's attributability and its contemporaneous, hash-manifested record resemble properties that audit regimes already demand of human work, though we claim compliance with no specific regime. A repository badge (increments audited, escalations raised, constitution version) would then do for machine science what CI badges did for software: it would compress an inspectable process into a glanceable, verifiable signal. None of this requires new technical infrastructure. It requires only that the artefact exist, which a conforming deployment produces.

\paragraph{A ratchet for standards.} The Constitution need not be static. Because rules are commits and receipts pin the version in force (\invariant{7}), standards can tighten over time without ambiguity about which version governed which increment. The safe dynamics have three channels. First, shadow-mode promotion: a candidate stricter rule enters as \texttt{ADVISORY}, rehearses enforcement on every cycle without gating, and accumulates ledger evidence on hit rate, dispute rate, and would-be blocking cost; the human promotes it to \texttt{BLOCKER} only when the record supports it. Second, threshold ratchets: numeric criteria tighten stepwise as the generator's competence grows, each step a commit with its rationale. Third, agent-proposed amendment: either agent may draft a rule change from what the ledger shows it. The loop can propose. Only the principal enacts. One rule matters most. Standards freeze within a cycle and move only between cycles, because an auditor that could raise the bar mid-round would face the generator with a moving target and dissolve the termination guarantee of \invariant{5}.

\paragraph{From copied glue to an installable protocol.} The reference implementation ships as copy-in glue: a deployer vendors \texttt{checks/} and \texttt{controller/} into an audit repository and edits three workflow files. The product line described in \S\ref{sec:productline} replaces that with an installable package whose command surface is the loop itself: scaffold a project, run the deterministic layer and the auditor against a pinned commit, verify a receipt, admit. The argument for packaging is more than convenience. Vendored copies drift, so two deployments can silently run different verifiers, and a receipt that pins the Constitution, the check sources, and the prompt, but not the machinery that verified and admitted on their basis, is bound one layer short of \invariant{7}'s intent. A release-versioned verifier closes it: the receipt can record the package version and distribution hash alongside the hashes it already carries, and the enforcement code becomes a pinned, citable dependency, not an unversioned copy. This binding is specified but not yet emitted, and it is the next thing we would fix. Packaging also gives field adaptation a natural unit, since check packs ship as plugins and DCL coverage can circulate between groups in the way we argue Constitutions should. The cost is a new trust surface, the package supply chain, for which hash-pinned installs and signed releases are the standard mitigations.

\paragraph{A console for the human principal.} The ledger is git-native on purpose, and it stays so. Raw git is nevertheless a demanding reading surface for the one human the protocol keeps in the loop, and \invariant{6} makes that human's asynchronous reading load-bearing. The product line therefore carries a supervision console over the ledger: cycle timelines, the advisory backlog by rule and hit rate, an escalation inbox assembling what an adjudication actually needs, and the telemetry a standards ratchet would read. One design rule carries protocol weight: the console writes nothing of its own. Every action it offers runs the same verbs the command line offers, so each materialises as a commit or an issue through the authenticated paths the agents already use, the ledger stays complete (\invariant{2}), and the console remains derivable from it. A supervision interface with a private database would accumulate the unversioned, unreplayable state the protocol exists to exclude. Three refusals follow from the same principle and are worth stating because a supervision display that breaks them is worse than none: it does not show a figure the ledger cannot support, it does not colour a step green before it happened, and it does not imply that it acted on its own. There is a boundary case that deserves its own honesty. Work in flight is not yet in the ledger, so progress reporting is a view, not a record: it lives in memory and dies with the process. The ledger holds every committed round but cannot know that a round was cut off, so an interrupted run must be marked as interrupted rather than left to read as finished.

\paragraph{Management by exception, at research scale.} As agent labour becomes cheap, the binding constraint of agentic science shifts from compute to human attention. CrossAudit's economic content is a hypothesised reduction in \emph{gating} cost from $O(\text{increments})$ to $O(\text{escalations})$, conditional on escalation volume staying low; no deployment here recorded the human attention spent, so the hypothesis is stated with its measurement plan, not as a result. Our own trial shows the failure mode plainly: the GPT-family arm's 10/10 false-block rate would make escalations $O(\text{increments})$. The economics hold only with a calibrated auditor and a decidably scoped Constitution, and they must be measured rather than assumed. This is the managerial precondition for one researcher credibly directing several agent pipelines at once. In principle it should extend beyond science to any agentic work product that arrives as discrete increments whose quality is partially rule-expressible, from code to quantitative analysis, a conjecture this paper does not test.

\section{Conclusion}\label{sec:conclusion}

CrossAudit reframes supervision of autonomous research from a platform feature into a public, versioned artefact produced by structurally independent reviewers. Its ingredients are plain: a second vendor, a rulebook with stable IDs, scripts that run before any model, a bounded loop, and git underneath. We chose them because a single researcher can adopt all of this without asking anyone's permission. The protocol's eight invariants are a deliberately small set. Together they buy structurally independent review, a re-inspectable history, and human authority at the boundary; everything else in this paper is reference detail, and the reference repositories are public. An AI scientist should not grade its own homework. With two repositories and two API keys, it no longer has to.

\paragraph{Author contributions and the use of language models.} This paper was written with heavy use of the class of tools it proposes to supervise, and the division deserves stating rather than implying. The protocol, the invariants, the study designs and every judgement about what the evidence supports are the authors'. A large language model drafted most of the paper's prose from the authors' outlines and notes, wrote the bulk of the reference implementation and the analysis scripts under the authors' direction, and produced the figures. The authors verified each of the reported numbers against the committed artefacts, checked all twenty-two references against their source pages, and read and revised the text. Two limitations of that arrangement should be named. First, the same assistant that drafted the prose also ran an internal screen of the manuscript for machine-authored writing patterns; the screen is committed under \texttt{audits/}, and it is a same-source review of the kind \S\ref{sec:related} argues against. Second, that screen missed a bibliographic error the assistant had itself introduced days earlier, which an external reviewer then caught. We report both because a paper about supervising machine work has no standing to be vague about the machine work in itself.


\end{document}